Review

# Scalable Causal Structure Learning: Scoping Review of Traditional and Deep Learning Algorithms and New Opportunities in Biomedicine

Pulakesh Upadhyaya[1,2], PhD; Kai Zhang[1], PhD; Can Li[1], MSPH; Xiaoqian Jiang[1], PhD; Yejin Kim[1], PhD

[1]School of Biomedical Informatics, University of Texas Health Science Center at Houston, HOUSTON, TX, United States
[2]Department of Biomedical Informatics, Emory University School of Medicine, Atlanta, GA, United States

**Corresponding Author:**
Pulakesh Upadhyaya, PhD
Department of Biomedical Informatics
Emory University School of Medicine
101 Woodruff Circle
Suite 4127
Atlanta, GA, 30322
United States
Phone: 1 9794225161
Email: pulakeshupadhyaya@gmail.com

## Abstract

**Background:** Causal structure learning refers to a process of identifying causal structures from observational data, and it can have multiple applications in biomedicine and health care.

**Objective:** This paper provides a practical review and tutorial on scalable causal structure learning models with examples of real-world data to help health care audiences understand and apply them.

**Methods:** We reviewed traditional (combinatorial and score-based) methods for causal structure discovery and machine learning–based schemes. Various traditional approaches have been studied to tackle this problem, the most important among these being the *Peter* Spirtes and *Clark* Glymour algorithms. This was followed by analyzing the literature on score-based methods, which are computationally faster. Owing to the continuous constraint on acyclicity, there are new deep learning approaches to the problem in addition to traditional and score-based methods. Such methods can also offer scalability, particularly when there is a large amount of data involving multiple variables. Using our own evaluation metrics and experiments on linear, nonlinear, and benchmark Sachs data, we aimed to highlight the various advantages and disadvantages associated with these methods for the health care community. We also highlighted recent developments in biomedicine where causal structure learning can be applied to discover structures such as gene networks, brain connectivity networks, and those in cancer epidemiology.

**Results:** We also compared the performance of traditional and machine learning–based algorithms for causal discovery over some benchmark data sets. Directed Acyclic Graph-Graph Neural Network has the lowest structural hamming distance (19) and false positive rate (0.13) based on the Sachs data set, whereas Greedy Equivalence Search and Max-Min Hill Climbing have the best false discovery rate (0.68) and true positive rate (0.56), respectively.

**Conclusions:** Machine learning–based approaches, including deep learning, have many advantages over traditional approaches, such as scalability, including a greater number of variables, and potentially being applied in a wide range of biomedical applications, such as genetics, if sufficient data are available. Furthermore, these models are more flexible than traditional models and are poised to positively affect many applications in the future.

## Introduction

**Background**

Many applications in biomedicine require the knowledge of the underlying causal relationship between various factors beyond association or correlation. Randomized controlled trials are widely used to uncover causality, but these experiments can be prohibitively expensive or unethical in many cases. Therefore, it has sparked an enormous amount of interest in identifying causal effects from observational data [1-3].

In this paper, we discuss causal structure learning; that is, learning causal relationships that are represented as directed graph structures between different factors and its application to biomedicine. The causal structure is represented by a causal graph (also called a causal Bayesian network), which is a directed acyclic graph (DAG), in which the nodes represent variables and edges represent causation (Figure 1). An edge is drawn from a variable that represents the cause to a variable that represents the effect of that cause. Based on a variety of methodologies, causal structure learning identifies which causal models represented by DAGs accurately represent the observed data.

For example, consider the example of a gene regulatory network [4-7], which is an abstract representation of the gene regulation processes as shown in Figure 1. By observing the data of multiple variables such as gene expression profiles, causal structure learning attempts to discover causal relationships among the genes. For example, if a gene A regulates another gene B, it is represented by an arrow between gene A and gene B.

Many researchers in the biomedical field are interested in causality and not just correlation (eg, whether a particular treatment affects a particular outcome). Unlike association- or correlation-based studies that simply indicate that any 2 variables are correlated, this approach seeks to determine the directional relationship between any 2 variables (eg, between a treatment variable and an outcome variable). In biomedicine, causal structure learning can be applied in a variety of applications.

**Figure 1.** Example of the causal structure. (A) A gene regulatory network is an abstracted structure (given by the directed graph on the right) of the complex biophysical process shown on the left. (B) A gene regulatory structure from the transmiR database for mice [6].

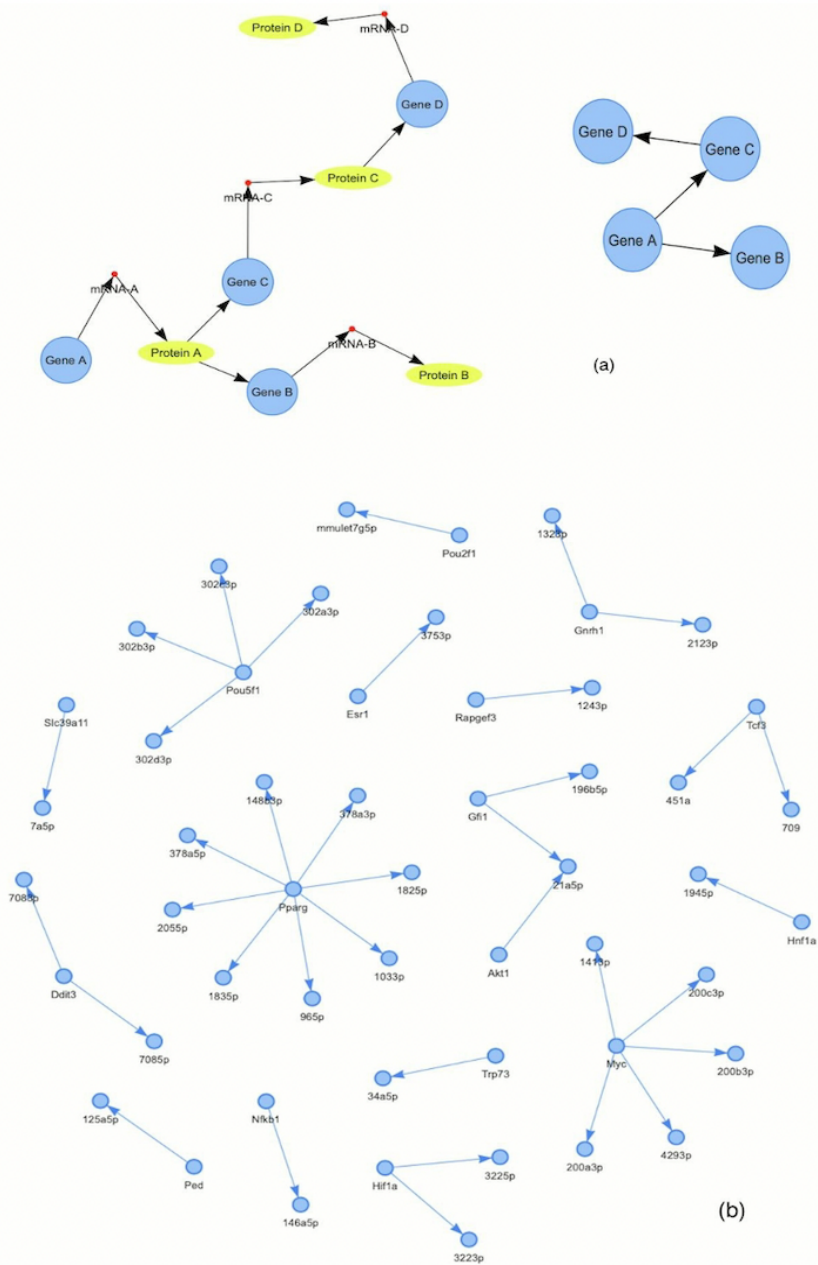

## Examples

### Gene Regulatory Networks

A gene regulatory network is a network in which molecular regulators and genes are the nodes, and the directed edges denote the interactions among them [5]. This is in contrast to association-based methods such as finding correlation or mutual information among the genes (finding Pearson, Kendall, Spearman correlation coefficients, etc) that do not have any directional information [8]. Such methods can only be accurate to a certain extent when it comes to deducing extensive gene regulatory structures from data sets with a large set of observations. Correlational studies can only indicate gene-gene association and not the direction of regulation. A gene regulatory network is an example of a causal structure that can be used to develop interventions to control gene expression.

Causal structure learning algorithms have been used to jointly deduce the phenotype network structure and directional genetic architecture [9]. It uses a difference causal inference method and compares it with another causal structure learning algorithm (difference-based Greedy Equivalence Search [GES]) as a baseline. Another study proposed a hybrid algorithm that combines Simulated Annealing with Greedy Algorithm to predict intergene transcriptional regulatory relationships [10], which are also directional in nature. In cancer, somatic genome alterations and differentially expressed genes have causal relationships. A correlational study cannot provide directional information in any of these applications.

The tumor-specific causal inference algorithm proposed by Xue et al [11] uses a Bayesian causal learning framework to find those relationships. Unlike association-based studies, this study is based on a causal structure learning framework across the whole genome where Ha et al [12] found gene signatures that were the causes of clinical outcomes and were not merely correlated to them. Apart from these examples, there are also networks such as those represented in the Sachs data set [13] that simultaneously incorporates measurements of 11 phosphorylated proteins and phospholipids to find causal pathways linking them. This is different from association-based correlation studies because protein signaling pathways are directional.

In our comparative analysis of the performance of this data set, we found that machine learning models can also be effective at finding causal structures (details are available in the *Results* section). In the case of more complicated protein signaling networks with many nodes, machine learning–based methods might be particularly effective.

### *Brain Connectivity Networks*

Different regions of the brain have distinct functions. Previous studies have used correlation-based methods [14] to find nondirectional functional connectivity among cortical regions. Spatial localization of brain functions has been studied using methods such as functional magnetic resonance imaging [15]. Regions within the brain are the nodes, and a directed edge between regions represents some functional connection (see Figure 2 in the paper by Brovelli et al [16] for the difference between coherence and causality graphs). Such connections are directional, can have different strengths (weights), and can be both inhibitory or excitatory [17]. Scalable causal structure learning models can also model such connection strengths in addition to directionality, which makes them more expressive than an association. In addition, brains have large-scale structural cortical networks that are directional with respect to information flow and can only be captured by causal structure instead of correlation.

**Figure 2.** Overview of the methods reviewed and benchmark results sections of the paper. CGNN: Causal Generative Neural Networks; DAG-GNN: Directed Acyclic Graph-Graph Neural Network; FCI: Fast Causal Interface; GAE: graph autoencoder; GES: Greedy Equivalence Search; GRAN-DAG: Gradient-based neural-directed acyclic graph learning; IC: inductive causation; LiNGAM: linear non-Gaussian acyclic model; MMHC: Max-Min Hill Climbing; PC: Peter Spirtes and Clark Glymour; RL-BIC: Reinforcement Learning-Bayesian Information Criterion; SAM: Structural Agnostic Modeling.

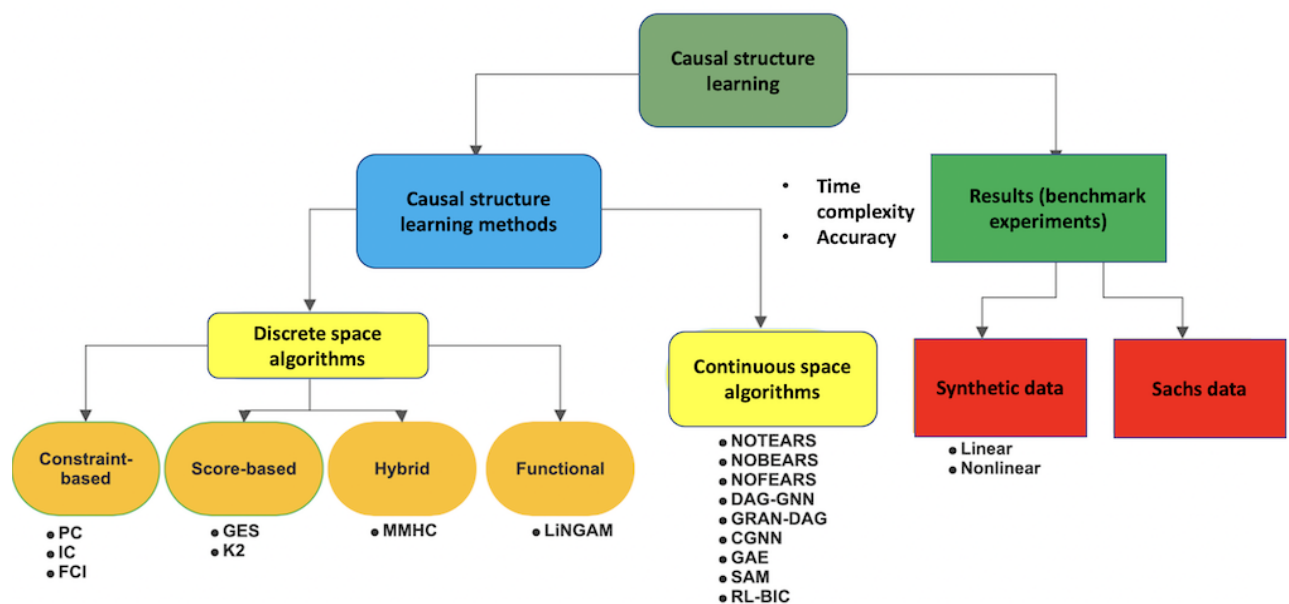

### *Epidemiology*

Causal structure learning has also been used in epidemiology with patients' medical records. Many complex diseases are multifactorial in which a combination of these factors contributes to disease predisposition. Causal structure learning considers multiple confounders to determine causal effects solely from one factor of interest to another. For example, causal structure has been used to disentangle psychological factors that predispose adolescents to smartphone addiction [18]. Incorporating a large set of medical claim records, a recent study used a scalable causal structure learning to elucidate the clinical pathways from comorbid illnesses to Alzheimer disease [19].

### **Challenges**

However, there are a few challenges. The general approach to solving this problem of learning a DAG from data, which has been studied for a long time [20], has a time complexity that scales exponentially with the number of observed variables. This is because the problem is generally nondeterministic polynomial-time complete [21]. In practice, if the number of variables is greater than a few hundred, the problem becomes intractable to solve optimally.

Several approaches have been used to solve this problem of intractable time complexity. Traditionally, constraint-based and score-based methods, which search for the optimal graph from a discrete space of candidate graphs, have been used to learn the DAG from data. Constraint-based methods such as the *Peter* Spirtes and *Clark* Glymour (PC) and Fast Causal Inference (FCI) algorithms (which will be discussed in detail in the *Discrete Space Algorithms* section) rely on statistical tests to estimate the correct causal structure. However, biological data usually involve hundreds to thousands of variables, and the complexity of algorithms increases exponentially as the number of variables increases. For example, typical human RNA sequence data contain at least 20,000 genes. Therefore, the complexity of the PC algorithm is proportional to $2^{20000}$, which is infeasible within a reasonable amount of time.

Hence, researchers have investigated various score-based methods that assign scores based on the data to each DAG and select the one with the best score. Although score-based methods scale better than constraint-based methods, they do not scale well for several thousand variables. On the other hand, patient medical records in electronic health records or claim data raise severe scalability concerns, because they include up to 144,000 International Classification of Diseases-Ninth Revision or 69,823 International Classification of Diseases-Tenth Revision diagnosis codes, >2000 US Food and Drug Administration–approved drugs, and >10,000 Current Procedural Terminology procedures or laboratory test codes.

To overcome the limited scalability of traditional methods, recent advances in machine learning algorithms have relaxed the problem of finding an optimal DAG into a continuous optimization problem with smooth acyclicity constraints. This enables the use of nonheuristic machine learning (including deep learning) algorithms to determine the optimal causal structure. This is a promising development in the field of biomedicine. In this study, we focus on scalable algorithms. Table 1 summarizes the algorithms discussed in this study. The tools available for some of these algorithms are listed in Multimedia Appendix 1. A list of ground truth causal structures can be found in the *bnlearn* repository [22].

There are 2 distinct approaches in the context of treatment effect evaluation: the structural approach and potential outcome framework approach [23]. In this study, we consider the first approach, in which there are 2 distinct types of algorithms for finding the causal DAG structure. In all of these examples, the goal is to learn a DAG that shows the directional relationship among variables from observational data.

**Table 1.** Summary of various algorithms for causal structure learning.

| Algorithm | DS[a] | CS[b] | Summary | Remarks | Scalability |
| --- | --- | --- | --- | --- | --- |
| PC[c] | ✓ | ✗ | A partially directed acyclic graph (CPDAG[d]) is produced by iteratively checking the conditional independence conditions of adjacent nodes, conditioned on an all-size subset of neighbors. | Outputs usually converge to the same equivalence class; high FPR[e] on experimental data | +[f] |
| IC[g] | ✓ | ✗ | Returns the equivalent class of the DAG[h] based on the estimated probability distribution of random variables and an underlying DAG structure. | Outputs usually converge to the same equivalence class. | + |
| FCI[i] | ✓ | ✗ | Modified PC algorithm to detect unknown confounding variables and produces asymptotically correct results. | Faster than PC with similar TPR[j]; converges to the same asymptotic result; high experimental FPR | ++ |
| GES[k] | ✓ | ✗ | Starts with an empty graph and iteratively adds and deletes edges in the graph by optimizing a score function. | Faster than PC with higher TPR; stable result for the same score function | ++ |
| Fast GES | ✓ | ✗ | Improved and parallelized version of GES | Faster than GES; same TPR; stable result for the same score function | ++ |
| K2 | ✓ | ✗ | Performs a greedy heuristic search for each nodes' parents. | Greedy searches might return very suboptimal solutions. | ++ |
| MMHC[l] | ✓ | ✗ | MMHC to find the skeleton of the network and constrained greedy search for edge orientation. | Greedy searches might return suboptimal solutions. | + |
| LiNGAM[m] | ✓ | ✗ | Transfer the linear structure model $x_i = \sum_{j<i} b_{ij} x_j + e_j$ to the form of $x = Bx + e$, and optimize for matrix B. | Works very well on linear data but not on nonlinear data. | ++ |
| NOTEARS | ✗ | ✓ | Uses smooth function $h(A)$, whose value characterizes the "DAG-ness" of the graph with adjacency matrix A—that is, $h(A)=0$ for DAG—and optimizes using continuous optimization. | Might converge to many different DAGs; GPUs[n] can speed up the process. | +++ |
| NOBEARS | ✗ | ✓ | Proposed a new acyclicity constraint that allows for faster optimization and scalability, and a polynomial regression loss to infer gene regulatory networks from nonlinear gene expressions. | Might converge to many different DAGs; GPUs can speed up the process. | +++ |
| DAG-GNN[o] | ✗ | ✓ | Uses an autoencoder framework and deep learning to train it and infer the causal structure from the weights of the trained network and is more scalable than NOTEARS. | Might converge to many different DAGs; GPUs can speed up the process. | ++++ |
| NOFEARS | ✗ | ✓ | Modify NOTEARS so the scoring function remains convex to ensure local minima. | Might converge to many different DAGs; GPUs can speed up the process. | ++++ |
| GAE[p] | ✗ | ✓ | Scalable graph autoencoder framework (GAE) whose training time increases linearly with the number of variable nodes. | Good accuracy; might converge to many different DAGs; GPUs can speed up the process. | ++++ |
| GRAN-DAG[q] | ✗ | ✓ | Extends the NOTEARS algorithm for nonlinear relationships. | Works on nonlinear data; better accuracy than NOTEARS; might converge to many different DAGs; GPUs can speed up the process. | ++++ |
| CGNN[r] | ✗ | ✓ | Generative model of the joint distribution of variables reducing MMD[s] between the graph and data. | Does not always converge to a single class of equivalent DAGs; GPUs can speed up the process. | ++++ |
| SAM[t] | ✗ | ✓ | Structurally agnostic model for causal discovery and penalized adversarial learning. | Does not always converge to a single class of equivalent DAGs; GPUs can speed up the process. | ++++ |
| RL-BIC[u] | ✗ | ✓ | Reinforcement learning-based algorithm that uses both the acyclicity constraint and the BIC[v] score. | Very good accuracy; does not always converge to a single class of equivalent DAGs; GPUs can speed up the process. | ++++ |

[a]DS: discrete space algorithms.

[b]CS: continuous space algorithms.

[c]PC: *Peter* Spirtes and *Clark* Glymour.

[d]CPDAG: completed partially directed acyclic graph.

[e]FPR: false positive rate.

[f]The + symbol for an algorithm indicates its scalability.

[g]IC: inductive causation.

[h]DAG: directed acyclic graph.

[i]FCI: Fast Causal Inference.

[j]TPR: true positive rate.

[k]GES: Greedy Equivalence Search.

[l]MMHC: Max-Min Hill Climbing.

[m]LiNGAM: linear non-Gaussian acyclic model.

[n]GPUs: graphical processing units.

[o]DAG-GNN: Directed Acyclic Graph-Graph Neural Network.

[p]GAE: graph autoencoder.

[q]GRAN-DAG: Gradient-based neural - directed acyclic graph learning.

[r]CGNN: causal generative neural network.

[s]MMD: maximum mean discrepancy.

[t]SAM: Structural Agnostic Modeling.

[u]RL-BIC: Reinforcement Learning-Bayesian Information Criterion.

[v]BIC: Bayesian information criterion.

### Paper Structure

This study attempts to provide a comparative study of various scalable algorithms that are used to discover causal structures from observational data to the biomedicine community. Some of these traditional and score-based methods have been extensively studied [24], but many of the algorithms discussed here focus on scalable causal structure learning. Although we do not list all possible approaches as Vowels et al [25], we sample a few important algorithms and evaluate their performance on synthetic data sets and the Sachs data set [13].

This tutorial paper presents algorithms for causal structure identification in biomedical informatics. In the *Methods* section, we discuss the methodology and examine the traditional algorithms that determine the optimal causal graph in a discrete space. We also discuss algorithms that use continuous space optimization to discover causal relationships. We compare the performance of these algorithms in the *Results* section. Finally, we present the discussion and conclusions. A brief overview of the methods and results is presented in Figure 2.

## Methods

### Overview

In this section, we discuss 2 paradigms of algorithms for causal structure learning. First, we consider algorithms that search for the optimal DAG in the discrete space of all possible DAGs (space of all possible discrete DAGs for a given number of variable nodes) or *discrete space algorithms*. Second, we consider scalable algorithms that use continuous optimization methods to find the optimal DAG (ie, algorithms that search the continuous space of all possible weighted DAGs to find the optimal one), known as *continuous space algorithms*.

### Discrete Space Algorithms

#### Overview

The first type that we discuss in this section is discrete space algorithms for causal discovery; that is, algorithms that search for the optimal DAG in the discrete space of candidate causal graphs. This is in contrast to continuous space algorithms (discussed in the *Continuous Space Algorithms* section) that search for the optimal DAG from the continuous space of weighted candidate graphs.

The discrete space algorithms can be divided into the following 4 types: combinatorial constraint-based models, score-based models, hybrid models, and functional models. In combinatorial constrained-based methods, we consider methods that check the conditional independence relations of 2 adjacent nodes conditioned on all subsets of their neighbors. Such methods can be useful when the number of variables is up to a few hundred. Score-based methods perform optimization by considering a score representing the goodness of fit and can handle more variables than constraint-based methods. Hybrid methods combine constraint- and score-based algorithms. Functional models find structural equations to describe the causal relationship and are useful mostly when the variables can be assumed to be expressed by some linear or nonlinear equations.

#### Combinatorial Methods

We now focus on combinatorial optimization methods, where conditional independence relationships in the data are used for finding the optimal DAG.

#### PC and Its Variants

The PC algorithm was proposed by *PC* and is named after them [26]. This algorithm produces a completed partially DAG (CPDAG) by iteratively checking the conditional independence relations of 2 adjacent nodes conditioned on all-size subsets of their neighbors. Three assumptions underlie the algorithm: no confounder variable, the causal Markov condition, and faithfulness. Under these conditions, this algorithm generates a partially directed causal graph that is proven to be asymptotically correct.

The PC algorithm is order-dependent; that is, the output of the algorithm can depend on the order in which the variables are provided to the algorithm. To address this problem, Colombo and Maathuis [27] developed a PC-stable algorithm in which the deletion of an edge takes place at the end of each stage (considering any 2 nodes' relations within a predetermined neighborhood). Thus, any ordering of vertices will result in the same edge deletions, resulting in the same stable output. The PCMCI and PCMCI+ [27-29] are 2 extensions of the PC algorithm proposed to handle large-scale time-series data sets.

**Inductive Causation Algorithm and Its Variants**

The inductive causation (IC) algorithm uses the estimated probability distribution of random variables with an underlying DAG structure and outputs the equivalent class of the DAG. In contrast, PC provides a schematic search method and is thus considered a refinement of the IC.

The IC* algorithm [30,31] is an extension of the IC algorithm, which searches for causal relations using observations of a set of variables, even when they appear as latent variables. The output of the IC algorithm is a CPDAG that only has directed edges (identified causation) and undirected edges (undetermined causation). The output of the IC* algorithm is an embedded pattern; that is, a hybrid graph containing ≥2 types of edges.

**FCI and Its Variants**

The FCI is a modification of the PC algorithm [30,32] that detects unknown confounding variables and produces asymptotically correct results. FCI improves the PC algorithm by adopting 2 rounds of phases of the PC algorithm. The algorithm first uses PC-phase I to find an initial skeleton, then uses the separation set to orient all v-structure triples (a->c<-b) and outputs a CPDAG; then performs another round of skeleton searching based on the CPDAG and repeats the orientation for unshielded triples. The really fast causal inference algorithm [33,34] skips the second step, which is the most time-consuming part of the task, and therefore significantly accelerates the FCI procedure. A set of 10 rules was added to the algorithm to orient the edges of the skeleton.

*Score-Based Methods*

**Overview**

In addition to traditional combinatorial methods such as PC and FCI, score-based methods have also been used to uncover causal structures. In these methods, algorithms determine the optimal DAG by optimizing a particular score.

A typical score function is the Bayesian information criterion (BIC) score. The GES algorithm uses different score functions for different data types as follows: the BIC score (for continuous data), likelihood-equivalence Bayesian Dirichlet uniform joint distribution score (for discrete data), and Conditional Gaussian score (for continuous or discrete mixture data).

$$BIC = k \ln(n) - 2 \ln(\hat{L})$$

where $\hat{L}$ is the maximized likelihood function of the model, *n* is the number of observational data points, and *k* is the degree of freedom. The definition of the Bayesian Dirichlet uniform joint distribution scoring function was found in a study by Buntine [35]. The conditional Gaussian score is defined on the ratios of joint distributions, and Andrews et al [36] have proved that the Conditional Gaussian score is *score equivalent*; that is, a scoring function that scores all DAGs in the same Markov Equivalence Class equally.

Score-based methods include the GES algorithm, the fast GES algorithm, and the K2 algorithm.

**GES Algorithm**

The GES algorithm was proposed by Chickering [37], and its underlying principles were obtained from Meek [37,38]. The algorithm starts with an empty graph and iteratively adds and deletes the edges in the graph by optimizing a score function. During the forward phase, the algorithm searches iteratively from the space of the DAGs created by one edge addition on the current DAG and selects the edge with the best score. The forward phase ends when the score is no longer increasing. In the second phase, the algorithm repeats the above step but deletes one edge at a time and selects the edge that improves the score the most. The algorithm stops as soon as there are no more edges to be deleted.

**Fast GES Algorithm**

Fast GES is an improved and parallelized version of the GES. Significant speedup was achieved by storing the score information during the GES algorithm [39]. In addition, several insights regarding parallelization were offered in the paper. First, the precalculation of covariances can be parallelized by variables. Second, it is possible to parallelize the process of calculating the edge scores when an edge addition is being performed on the graph. A greater speedup can be achieved for sparse graphs.

**K2 Algorithm**

The main idea of the K2 algorithm [40] is to perform a greedy heuristic search of the parents of each node. For each node, the algorithm iteratively determines the parents. When visiting node $X_i$, the algorithm searches for all possible parents of $X_i$ ($X_j$ such that *j* has a lower ordering of *i*). The algorithm greedily adds $X_j$ to the parent set of $X_i$ if it could increase a predefined score function. The iteration for node $X_i$ stops when the number of parent nodes reaches the (preset) maximum or when adding an $X_j$ does not increase the score anymore. The entire algorithm finishes after completing the iteration for all $X_i$.

*Hybrid Algorithms*

Hybrid algorithms use a combination of score-based and combinatorial constraint-based optimization methods to determine the optimal DAG. An example is the Max-Min Hill Climbing (MMHC) algorithm. The MMHC algorithm is a combination of constraint- and score-based algorithms [41]. It uses the Max-Min Parents and Children algorithm. A detailed description is provided in [41,42] to find the skeleton of the Bayesian network and then perform the constrained greedy search to orient the edges.

### Algorithms for Functional Causal Models

#### Overview

Functional causal models or structural equation models (SEMs) assume structural equations that define the causal relationships. Such structural equations may describe the linear and nonlinear relationships among variables. In addition to the discrete methods discussed here, SEMs are also an important assumption in many machine learning–based methods that use the continuous optimization techniques in the *Continuous Space Algorithms* section.

#### Linear Non-Gaussian Acyclic Model

The linear non-Gaussian acyclic model (LiNGAM) was originally proposed by Shimizu [43] to learn linear non-Gaussian acyclic causal graphs from continuous-valued data. The LiNGAM transfers $x_i = \sum_{i<j} b_{ij} x_j + e_i$ to the form of $x = Bx + e$, and the causal structure problem becomes an optimization problem for matrix $B$. There are several extensions of the LiNGAM model using different estimation methods, including independent component analysis–based LiNGAM [43,44], DirectLiNGAM [45], and Pairwise LiNGAM [46].

#### Additive Noise Models

A nonlinear additive noise model is proposed in [47]. The model assumes that the observed data are generated according to the following equation:

$$x_i = f_i(x_{pa(i)}) + n_i$$

where $f_i$ is an arbitrary function, $x_{pa(i)}$ denotes the ancestor nodes of node $x_i$ in the true causal graph, and $n_i$ is the noise variable of an arbitrary probability density function. This study proves the basic identifiability principle for the 2 variables case and generalizes the results to multiple variables.

### Continuous Space Algorithms

#### Overview

Traditional causal discovery algorithms attempt to discover a causal graph, which is usually a DAG, while searching for an optimal graph in the space of candidate graphs. The score-based optimization problem of DAG learning (discussed in the *Score-Based Methods* section) is mathematically given by the following equation:

$$\min_{A \in R^{d \times d}} F(A), \text{ where } G(A) \in \Delta$$

Here, $\Delta$ is the set of all DAGs with $d$ nodes, and $F(A)$ is the cost or score function. The problem of searching for all DAGs is usually intractable and superexponential in the number of nodes in the graph.

An alternative approach would be to model the problem as a continuous space optimization problem, which would then allow the application of various learning techniques. Recently, several publications have explored continuous optimization methods that learn DAGs by adding an acyclicity constraint. In these approaches, the discrete acyclicity constraint $G(A) \in \Delta$ is replaced by $h(A) = 0$, where $h(A)$ is a smooth function that ensures acyclicity of $G(A)$.

The hard constraints on acyclicity can be relaxed and incorporated into the loss function to be optimized. This smooth continuous constraint allows the use of machine learning–based tools, which in turn can make the algorithms scalable in the presence of substantial amounts of data. These algorithms are based on SEMs.

#### NOBEARS Algorithm

Several other improvements such as the NOBEARS algorithm [48] have improved the scalability of the NOTEARS algorithm. A fast approximation of a new constraint is proposed, and a polynomial regression loss model is proposed to account for nonlinearity in gene expression to infer gene regulatory networks.

#### NOTEARS Algorithm

This algorithm considers the acyclicity constraint and comes up with the constraint.

$$h(A) = Trace[(exp(A \circ A)] - d.$$

Here, $\circ$ is the element-wise product. $h(A)$ equals 0 if and only if $G(A)$ is acyclic, and more severe deviations from acyclicity would increase the value of the function. This study assumes a linear SEM:

$$X_i = A^T X_i + Z_i$$

Here, $X_i$ is a $d$-dimensional sample vector of the joint distribution of $d$ variables and $Z_i$ is a $d$-dimensional noise vector. We denote $n$ such samples by matrix $X$, and the loss function (with $l_1$-regularization) is given as follows:

$$F(X, W) = \frac{1}{2n} \| X - AX \|_F + \lambda \| A \|_1$$

The constraint is given by $h(A)=0$ and is used in the final Lagrangian formulation of the loss function. The paper on learning sparse nonparametric DAGs is an extension of NOTEARS, which tries to define a "surrogate" of the matrix above for general nonparametric models to optimize [49].

### Directed Acyclic Graph-Graph Neural Network Algorithm

A Directed Acyclic Graph-Graph Neural Network (DAG-GNN) [50] generalizes the NOTEARS algorithm by considering the nonlinearity in the SEMs. It can be modeled with a variational autoencoder neural network with a special structure, with an encoder $Z = g_1((I - A^T)^{-1} g_2(X))$, and a decoder $X = f_2((I - A^T)^{-1} f_1(Z))$ and where $g_1, g_2$ are parameterized functions that can be assumed to serve as the inverse of $f_1, f_2$, respectively.

This variational framework considers $Z$ to be a latent vector (instead of viewing it as noise in linear SEMs), which can have dimensions other than $d$. The decoder then attempts to reconstruct the data from this latent variable. The encoder and decoder can be trained together from $n$ samples of $X, (X_1, X_2, ..., X_n)$ such that the loss function:

$$F(X, A, \theta) = \frac{1}{n} \sum_{k=1}^{n} [\log p(X_k|Z)] - KLD(q(Z|X_k) \| p(Z))$$

is minimized, where *KLD* is the Kullback-Liebler Divergence. The constraint in this optimization process to ensure the acyclicity of matrix *A* is slightly modified to:

$$h(A) = Trace[(I_d + \alpha A \circ A)^d] - d = 0$$

where α is an arbitrary parameter. This constraint can be implemented more easily in graphical processing unit-based deep learning libraries owing to the algorithm's parallelizability and scalability of the algorithm.

### NOFEARS Algorithm

Wei et al [51] demonstrated that the NOTEARS algorithm fails to satisfy the Karush-Kuhn-Tucker regularity conditions. Therefore, they reformulated the problem to ensure that the convexity of the scoring function can still ensure local minima even when the constraints are nonconvex. This new algorithm called the NOFEARS algorithm has the following acyclicity constraint.

$$h(A) = \sum_{p=1}^{d} c_p Trace[A^p] = 0$$

### Graph Autoencoder

Ng et al [52] propose another graph autoencoder (GAE) framework for causal structure learning, which improves the training speed and performance over DAG-GNN for both linear and nonlinear synthetic data sets.

Some other similar machine learning–based continuous learning algorithms include gradient-based neural DAG [53], Causal Generative Neural Network [54], and structurally agnostic model [55].

### Reinforcement Learning-Based Methods

Reinforcement learning-based methods have been proposed recently that consider both the acyclicity constraint and BIC score in the reward function and attempt to learn the DAG [56]. They used an actor-critic model, where the actor is an encoder-decoder framework that takes data as input and outputs the graph. The critic uses the reward function for this graph and updates the proposed graph.

## Results

This section provides the results to compare the effectiveness of some causal structure learning algorithms on synthetic and real data.

**Benchmark Methods**

The synthetic data were generated in the same manner as in the DAG-GNN paper [50]. An Erdos-Renyi model with an expected node degree of 3 was used to generate the random graph, and the adjacency matrix was formed by assigning weights to the edges from a uniform distribution. The samples were generated using the following structural equation:

$$X = g(A, Z, X)$$

Here, $Z$ is random Gaussian noise. We consider 2 functions for $g(X)$. The first is a (linear) identity function:

$$g(A, Z, X) = A^T X + Z$$

and the second is a nonlinear function

$$g(A, Z, X) = A^T \cos(X + 1) + Z$$

We considered 5 data sets for both linear and nonlinear functions. For each data set, we generated n=5000 independent samples according to the above equations. We used 6 algorithms, 4 of which are discrete space algorithms. PC and Greedy Fast Causal Interface (GFCI) are constraint-based methods, GES is a score-based method, and MMHC is a hybrid method. We also considered 2 continuous space methods, DAG-GNN and GAE.

We also evaluated these algorithms on the publicly available Sachs data set [13] using the above 4 metrics and showed the results are shown in Table 2. For other data sets that have the ground truth but are not covered in our experiments, please refer to the *bnlearn* repository [22]. The algorithms were implemented in Python for machine learning–based continuous space methods and R for discrete space algorithms.

**Table 2.** Benchmark experiments on the Sachs data set. We evaluated 6 algorithms— Peter Spirtes and Clark Glymour (PC), Greedy Equivalence Search (GES), Greedy Fast Causal Interface (GFCI), Max-Min Hill Climbing (MMHC), Directed Acyclic Graph-Graph Neural Network (DAG-GNN), and graph auto encoder (GAE), on 4 metrics—structural hamming distance (SHD), true positive rate (TPR), false positive rate (FPR), and false discovery rate (FDR)—and show their results in Figure 3. In all these evaluations, we consider any edge whose direction is reversed as half discovered.

| Metric | PC | GES | GFCI | MMHC | DAG-GNN | GAE |
|---|---|---|---|---|---|---|
| SHD (↓) | 24.50 | 26.50 | 29.50 | 22.00 | *19.00* [a] | 22.00 |
| FDR (↓) | 0.77 | 0.72 | 0.79 | *0.68* | 0.71 | 0.89 |
| TPR (↑) | 0.32 | *0.56* | 0.44 | 0.47 | 0.11 | 0.05 |
| FPR (↓) | 0.49 | 0.64 | 0.72 | 0.45 | *0.13* | 0.21 |

[a]Italicized values represent the best results for each metric.

**Figure 3.** Accuracy comparison. We evaluated 6 algorithms: Peter Spirtes and Clark Glymour (PC), Greedy Equivalence Search (GES), Greedy Fast Causal Interface (GFCI), Max-Min Hill Climbing (MMHC), Directed Acyclic Graph-Graph Neural Network (DAG-GNN), and graph auto encoder (GAE) on 4 metrics, structural hamming distance (SHD↓), true positive rate (TPR↑), false positive rate (FPR↓), and false discovery rate (FDR↓). In all these evaluations, we considered any bidirectional edges as half discovered. In experiments (A-D), first column, the data are drawn from a distribution according to the underlying causal graph where relationships between nodes are linear, and experiments (E-H), second column, is for the nonlinear case. In all experiments, the number of nodes of the graph ranges from 10, 20, 50, to 100. For each graph size, we drew 5 different data sets from the graph structure with a sample size of 1000 and calculated 4 evaluation metrics and obtained the average.

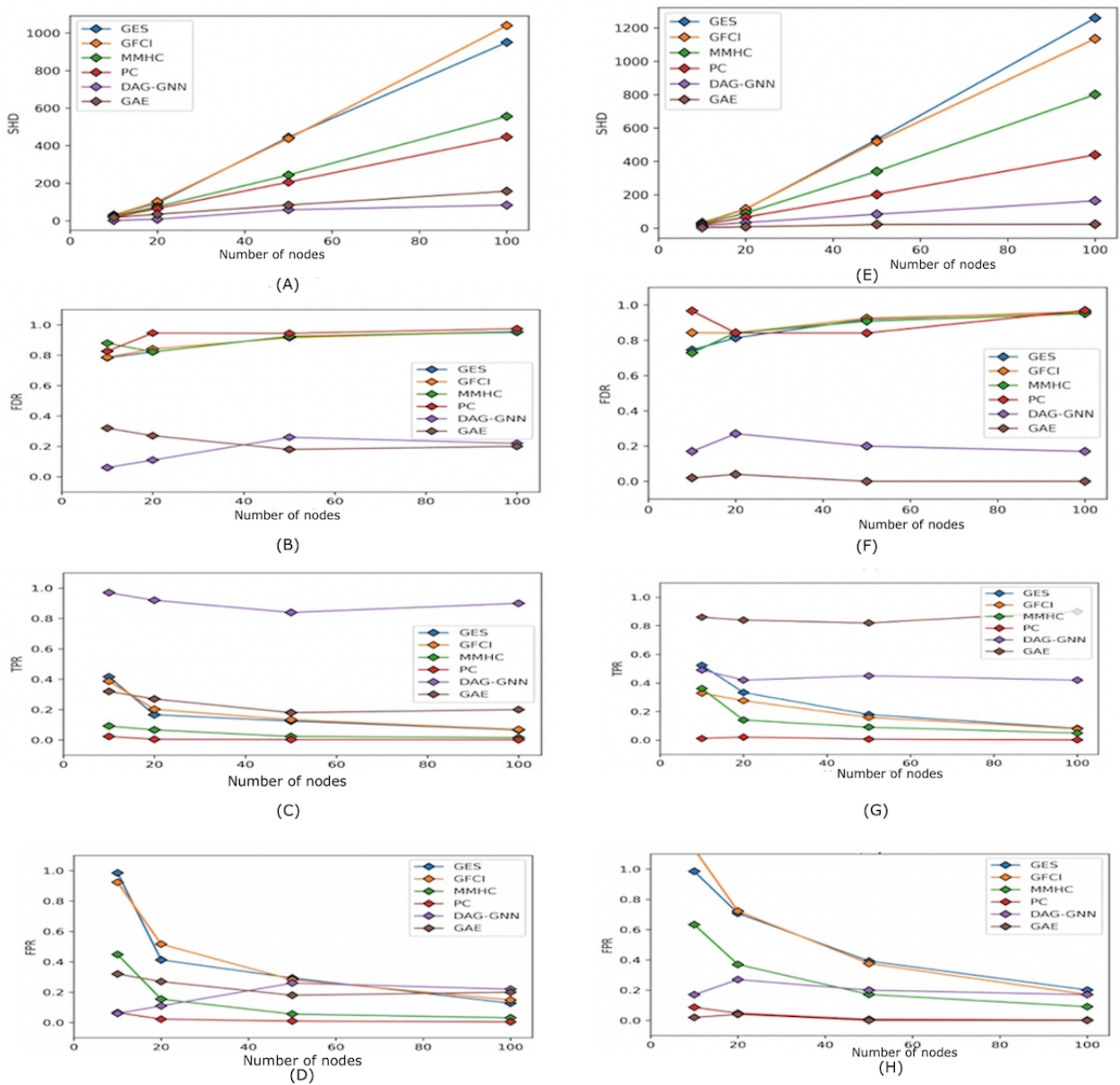

## Observations

All algorithms were tested on both linear and nonlinear data. The accuracies of some of these algorithms are shown in Figure 3. We used the following 4 evaluation measures: structural hamming distance (SHD), true positive rate (TPR), false positive rate (FPR), and false discovery rate (FDR). SHD refers to the number of edge insertions, deletions, and reversals. In our case, we used a modified SHD, where a reversal contributes half of the SHD score instead of 1. The TPR is the ratio of the algorithm's correctly discovered edges to the number of edges in the ground truth graph. FPR is the ratio of the algorithm's falsely discovered edges to the number of nonedges in the ground truth graph. FDR is the ratio of the algorithm's falsely discovered edges to the total number of discovered edges. We also evaluated these algorithms on the Sachs data set using the above 4 metrics, and the results are shown in Table 2.

## Time Complexity

The relative scalability of different algorithms is presented in Table 1. The worst-case time complexity of the PC, GES, and GFCI algorithms was $O(2^m)$, where $m$ is the number of variables (nodes in the DAG). For the GES, the best-case time complexity was $O(m^2)$. For the GAE and DAG-GNN, the time complexity of the algorithm is $O(km^2)$, where $k$ is the number of iterations. The time complexity for MMHC is $O(|V|^2 |S|^{l+1})$, where $V$ is the set of variables, $S$ is the largest set of parents and children, and $l$ is a parameter of the algorithm that denotes the size of the largest conditioned subset [41].

In our experiments, we observed that in the worst-case scenario, the running time for a maximum of 100 variables was of the order of hours for MMHC and of the order of minutes for the other algorithms. However, as the number of variables increases to a few thousand, machine learning–based methods such as DAG-GNN and GAE can provide solutions in a reasonable time. The trade-off between complexity (number of iterations) and accuracy can provide a choice between a method that is less accurate but faster or vice versa.

## *Discussion*

### Interpretation of Results

It is clear from the results that the algorithms have different advantages and disadvantages. Although the PC algorithm performs well across both linear and nonlinear data, it has a low TPR and is computationally intensive. The GES, GFCI, and MMHC algorithms show a very high FPR, but their TPR is higher than that of the PC algorithm. The SHD of the 2 machine learning–based methods—DAG-GNN and GAE—was also considerably lower for both data sets.

Continuous constraint-based algorithms generally exhibit a very low FDR, except for the benchmark Sachs data set. This is generally because both linear and nonlinear models are based on SEMs with the same causal relationship function at every node, which is an algorithm assumption when they learn the causal structure, but one cannot guarantee the same for Sachs data [13], because such constraints cannot be defined a priori.

This is corroborated by recent results from Zhu et al [56] where such gradient-based methods performed poorly on data generated by a nonlinear model, in which every causal relationship (node function) was sampled from a Gaussian distribution. However, this is a growing research area. In general, in areas such as gene regulatory networks and brain connectivity networks where the number of variables is large, machine learning–based methods can provide comparable results to traditional methods with a much more efficient time complexity and scalability.

### Challenges

Machine learning for causal structure learning is not without its limitations, which may present several challenges. First, in many applications, there is no ground truth about causal structure, which makes it difficult to evaluate the performance of these algorithms. Furthermore, many scalable methods use stochastic gradient descent; thus, the final output graph is not always deterministic. When the number of data samples or variables is low, traditional or score-based methods are a better choice, especially when the application requires fewer false positives. For the PC, GES, and GFCI algorithms, we observed that these algorithms require considerable running time, as the number of variables is more than 100 [57].

However, when it comes to large samples of data (eg, more than 100,000 samples) or hundreds of variables (eg, in many gene networks), machine learning methods can provide a reasonable solution, because other methods fail owing to scalability issues. As machine learning algorithms are highly parallelizable, the solutions can be computed much faster, particularly through the use of a graphical processing unit. These algorithms are potentially useful for many applications related to genetics and biomedicine, especially those with an abundance of observational data.

The continuous space machine learning models are more scalable and might be useful in the era of big data. Traditional methods might have complexities that grow exponentially with the number of attributes. Despite the nonconvexity of the optimization proposed by Zheng et al [58], optimization and learning strategies can be used to help find the optimal solution. Several methods have been used to solve this problem using augmented Lagrangian approaches [50,52].

The NOBEARS algorithm reduces the computing complexity of NOTEARS from cubic to quadratic in terms of the number of attributes [48], allowing for smooth implementation in data sets that have more than 4000 attributes. The algorithms are also highly parallelizable, and most of the algorithms use deep learning libraries such as Tensorflow [59] and PyTorch [60].

Machine learning techniques for causal discovery, which use continuous space optimization, are an emerging area of research, which can lead to more efficient causal discovery, particularly in applications where directed graphs are used to specify causal relations more clearly. With sufficient data, machine learning models can be robust to certain discrepancies such as sample bias, missing data, and erroneous measurements. Many of these applications have also focused on weaker concepts of causality such as pairwise directionality during the analysis of gene networks and brain connectivity networks [61,62].

It is noteworthy that machine learning methods are usually black box methods, which might provide lesser insight into the process of derivability of the causal structures. For higher interpretability, an option that has been explored is to develop parallel versions of these algorithms, such as PC [63]. In the future, options such as ensemble learning can be explored for the same.

Some other challenges can be found in finding causal structure from data. In the case of learning causal structure from electronic health record data, they might have several problems, such as missing values or noise in the data, which are very common [64]. If the number of missing values or the amount of noise is significant, the application of causal discovery methods might yield unreliable results.

Furthermore, most causal discovery methods assume that the distribution of data is stationary, which may not be true in certain medical applications [65]. Hence, it is very important to consider the aforementioned problems as well as issues related to selection bias before causal structure learning methods are applied. Glymour et al [24] discuss some general guidelines to avoid such problems in causal structure learning. These generalized learning algorithms are ineffective in many biomedical applications, such as in learning biological or gene networks, because they do not consider specific network constraints. These constraints can be incorporated into causal structure learning methods for greater efficiency.

## Conclusions

In this paper, we have discussed the motivation for causal structure discovery in biomedicine as well as some interesting applications. Two paradigms of causal discovery algorithms have been reviewed. Combinatorial or score-based algorithms are used in the first paradigm for optimizing discrete spaces of candidate causal graphs, whereas machine learning algorithms are used in the second paradigm to solve continuous optimization problems with acyclicity constraints. In addition to listing these methods, we have also included resources that readers can use to find appropriate applications. Furthermore, we tested several algorithms against synthetic benchmark data sets and against the Sachs real-world data set and evaluated their relative performances. We have also discussed their theoretical time complexity. Our discussion of the limitations and challenges of various algorithms is intended to offer readers a guide for choosing an algorithm from among the many available options. Finally, we highlight several challenges associated with finding causal structure from real-world data (eg, missing values, nonstationarity, noise, and sampling bias).

## Acknowledgments

XJ is a Cancer Prevention and Research Institute of Texas scholar in cancer research (RR180012) and was supported in part by the Christopher Sarofim Family Professorship, University of Texas Stars award, University of Texas Health Science Center startup, and the National Institutes of Health under award numbers R01AG066749 and U01TR002062.

## Authors' Contributions

The survey and experiments on deep learning–based methods and the survey on potential applications were conducted by PU. The survey and experiments on traditional methods were conducted by KZ and CL. XJ and YK conceived the study and provided useful inputs for the potential applications of scalable structure learning.

## Conflicts of Interest

None declared.

## Multimedia Appendix 1

List of tools for causal discovery.
[[DOCX File , 21 KB](#)-[Multimedia Appendix 1](#)]

XSL•FO
RenderX

## Abbreviations

**CPDAG:** completed partially directed acyclic graph
**DAG:** directed acyclic graph
**DAG-GNN:** Directed Acyclic Graph-Graph Neural Network
**FCI:** Fast Causal Inference
**FDR:** false discovery rate
**FPR:** false positive rate
**GAE:** graph autoencoder
**GES:** Greedy Equivalence Search
**GFCI:** Greedy Fast Causal Interface
**IC:** inductive causation
**LiNGAM:** linear non-Gaussian acyclic model
**MMHC:** Max-Min Hill Climbing
**PC:** Peter Spirtes and Clark Glymour
**SEM:** structural equation model
**SHD:** structural hamming distance
**TPR:** true positive rate